\documentclass[runningheads]{llncs}
\usepackage{graphicx}
\usepackage{mathtools}
\usepackage{todonotes}
\usepackage{hyperref}
\usepackage{amsmath,amssymb}
\usepackage{appendix}
\usepackage{arydshln}

\usepackage{subcaption}
\usepackage{mwe}

\newcommand{\ubold}[1]{\fontseries{b}\selectfont#1}
\usepackage{fixmath}
\usepackage{bm}

\newcommand{\VV}{\mathbb{V}}
\newcommand{\EE}{\mathbb{E}}

\DeclarePairedDelimiterX{\infdivx}[2]{[}{]}{%
  #1\;\delimsize\|\;#2%
}
\newcommand{\infdiv}{\text{KL}\infdivx}

\DeclareMathOperator{\E}{\mathbb{E}}

\begin{document}
\title{Supervised Uncertainty Quantification for Segmentation with Multiple Annotations}
\titlerunning{Supervised Uncertainty Quantification for Segmentation}
\author{Shi Hu\inst{1} \and Daniel Worrall\inst{1} \and Stefan Knegt\inst{1} \and Bas Veeling\inst{1} \and Henkjan Huisman\inst{2} \and Max Welling\inst{1}}
% index{Hu, Shi}
% index{Worrall, Daniel}
% index{Knegt, Stefan}
% index{Veeling, Bas}
% index{Huisman, Henkjan}
% index{Welling, Max}

%\author{First Author\inst{1}\orcidID{0000-1111-2222-3333} \and Second Author\inst{2,3}\orcidID{1111-2222-3333-4444} \and Third Author\inst{3}\orcidID{2222--3333-4444-5555}}
%
\authorrunning{S. Hu et al.}
\institute{University of Amsterdam \and Radboud University Medical Center}
\maketitle              % typeset the header of the contribution
\begin{abstract}
The accurate estimation of predictive uncertainty carries importance in medical scenarios such as lung node segmentation. Unfortunately, most existing works on predictive uncertainty do not return \emph{calibrated} uncertainty estimates, which could be used in practice. In this work we exploit multi-grader annotation variability as a source of `groundtruth' aleatoric uncertainty, which can be treated as a target in a supervised learning problem. We combine this groundtruth uncertainty with a Probabilistic U-Net and test on the LIDC-IDRI lung nodule CT dataset and MICCAI2012 prostate MRI dataset. We find that we are able to improve predictive uncertainty estimates. We also find that we can improve sample accuracy and sample diversity. In real-world applications, our method could inform doctors about the confidence of the segmentation results.

\end{abstract}

\section{Introduction}
In recent years, deep learning has propelled the state of the art in segmentation in medical imaging \cite{GuLYZYZGWZ18,GruetzemacherGP18,Causeyetal2018,WangZLLGZDGT17,kohl2018probabilistic}. However, previous works tend to focus on maximizing accuracy, ignoring predictive uncertainty. Modeling uncertainty at the per-pixel level is as important as accuracy, especially in medical scenarios, since it informs clinicians about the trustworthiness of a model's outputs \cite{biqt,kendallgal}.

Typically, there are two main types of uncertainty one cares about, \emph{aleatoric} and \emph{epistemic} \cite{aleaorepi}. Aleatoric uncertainty is a measure of the intrinsic, irreducible noise found in data, usually associated with the data acquisition process. Epistemic uncertainty is our uncertainty over the true values of a model's parameters, which arises from the finite size of training sets. With increasing training set size, epistemic uncertainty tends asymptotically to zero \cite{galthesis}. In practice, these two sources of uncertainty are difficult to quantify. Typically epistemic uncertainty is very hard to evaluate since one would need access to the groundtruth model to measure it, but it is possible to form a meaningful estimate of aleatoric uncertainty since we do have access to groundtruth data. Consider a training set of $N$ images $\{x_i\}_{i=1}^N$. If for the $i$\textsuperscript{th} image we are able to acquire $D$ grader segmentations $\{y_i^1,...,y_i^D\}$, then we define aleatoric uncertainty to be the per-pixel variance among these segmentations $\VV_{p(\mathcal{D})}[y_i] = \frac{1}{D}\sum_{j=1}^D (y_i^j - \bar{y}_i)^2$, where $\bar{y}_i = \frac{1}{D}\sum_{j=1}^D y_i^j$. Datasets containing multiple annotations exist in the literature, such as \cite{lidc1,lidc2,lidc3,da16}, and it is surprising that to date the authors cannot find examples of intergrader variability being exploited.

In this paper, we build a segmentation model based on the Probabilistic U-Net \cite{kohl2018probabilistic}, exploiting intergrader variability as a target for aleatoric uncertainty. With this model one can draw diverse segmentations from its output and gain quantitative, calibrated aleatoric uncertainty estimates. We further add a source of epistemic uncertainty, which the model previously did not have. To view these two uncertainties we deploy an uncertainty decomposition in the output-space based on the law of total variance. We find improved predictive performance as well as better aleatoric uncertainty estimation over previous works, while also achieving higher sample diversity, which we did not explicitly design in. 

\section{Background And Related Works}
Below we provide an overview of predictive uncertainty for deep learning.

\noindent\textbf{Predictive Uncertainty}
\sloppy Consider a training set $\mathcal{D} = \{(x_i, y_i)\}_{i=1}^N$ with inputs $x_i$ and target segmentations $y_i$, and a neural network with parameters/hidden variables $\theta$. We can think of the neural network as a conditional distribution $p(y|x,\theta)$. Given test image $x_*$, the posterior predictive distribution \cite{MacKay1992} is $p(y_* | x_*, \mathcal{D}) = \int p(y_* | x_*, \theta) p(\theta | \mathcal{D}) \, \mathrm{d}\theta$, where $p(\theta|\mathcal{D})$ is a posterior distribution over $\theta$ given the training data. This quantity is intractable to find \cite{BleiKM16}, so it is typically approximated by some $q_\lambda(\theta)$ from a tractable family of distributions, where $\lambda$ is called the \emph{variational parameters}. Typically the approximation is fitted by minimizing the reverse KL-divergence $\infdiv{q_\lambda(\theta)}{p(\theta | \mathcal{D})}$. This is intractable, since it contains the intractable posterior term, but can be rearranged into the ELBO \emph{evidence lower-bound} \cite{BleiKM16}:

\begin{align}
    \min_\lambda \infdiv{q_\lambda(\theta)}{p(\theta | \mathcal{D})} = \max_\lambda \mathbb{E}_{q_\lambda(\theta)}[\log p(\mathcal{D} | \theta)] - \infdiv{q_\lambda(\theta)}{p(\theta)} 
\end{align}

\noindent where $p(\theta)$ is a prior on $\theta$ and $p(\mathcal{D} | \theta) = \prod_{i=1}^N p(y_i | x_i, \theta)$. The \emph{predictive uncertainty} is the variance of the posterior predictive distribution.

\noindent\textbf{Aleatoric and Epistemic Uncertainty}
The predictive uncertainty can be decomposed into two parts. By the law of total variance, we can write predictive variances as a sum of these two independent components:
\begin{align}
    \underbrace{\VV_{p(y|x)}[y]}_\text{predictive uncertainty} =  \underbrace{\E_{q_\lambda(\theta)}[\VV_{p(y|\theta,x)}[y]]}_\text{aleatoric uncertainty} + \underbrace{\VV_{q_\lambda(\theta)}[\E_{p(y|\theta,x)}[y]]}_\text{epistemic uncertainty},
\end{align}
where we have used the notation $\mathbb{E}$ and $\VV$ for the expectation and variance operator. We have labeled the two right-hand terms as \emph{aleatoric} and \emph{epistemic} uncertainty. The aleatoric term measures the average of the output variance $\VV_{p(y|\theta,x)}[y]$, under all settings of the variables $\theta$. If $q_\lambda(\theta)$ were a delta peak, we would expect this term not to vanish and thus is it associated with aleatoric (data) uncertainty \cite{biqt}. The epistemic term measures fluctuations in the mean prediction. These fluctuations exist because of uncertainty in the approximate posterior $q_\lambda(\theta)$. If $q_\lambda(\theta)$ were a delta peak, then this term would vanish to zero, and thus we associate it with epistemic (model) uncertainty \cite{biqt,kendallgal}.

Current techniques for estimating aleatoric and epistemic uncertainty follow similar line. In Tanno \emph{et al.} \cite{biqt} the authors treat MRI superresolution as a regression problem. They build a CNN directly outputting $\mathbb{E}_{p(y|\theta,x)}[y]$ and $\VV_{p(y|\theta,x)}[y]$. They model epistemic uncertainty using variational dropout \cite{vdp}. Bragman \emph{et al.} \cite{Bragman2018} build on this technique, applying it to radiotherapy-treatment planning and multi-task learning. Concurrent to \cite{biqt} Kendall and Gal proposed a similar method using Monte Carlo (MC) instead of variational dropout \cite{mcdropout}.They also proposed a method which would work for classification, where they predict a mean and variance in the logit-space just before a sigmoid. Jungo \emph{et al.} \cite{Jungo2018} estimate epistemic uncertainty in the context of postoperative brain tumor cavity segmentation using MC dropout \cite{mcdropout}. In \cite{Ayhan2018} Ayhan and Berens treat the data augmentation process as part of the approximate posterior $q_\lambda(\theta)$. They claim this is aleatoric uncertainty, but from their method it appears they really compute epistemic uncertainty. None of these works quantitatively evaluates the quality of the epistemic and aleatoric uncertainties. In this work, we show that the aleatoric uncertainty can indeed be measured.

\noindent\textbf{The Probabilistic U-Net}
In the Probabilistic U-Net \cite{kohl2018probabilistic}, the approximate posterior distribution is given the form $q_\lambda(z|x,y)$, where we have set $\theta=z$. The hidden variables are thus activations $z|x,y$ dependent on the training data. A (conditional) prior over $z$ is given by a \emph{prior network} $p_\lambda(z|x)$. To train this setup, the authors employ a variant of the ELBO with a $\beta$-weight on the KL-penalty
\begin{align}
    \max_\lambda \frac{1}{N}\sum_{i=1}^N \left (\E_{q_\lambda(z| x_i, y_i)} \left [\log p_\lambda(y_i | x_i, z) \right ] - \beta \cdot \infdiv{q_\lambda(z | x_i, y_i)}{p_\lambda(z | x_i)} \right). \label{eq:probunetloss}
\end{align}
Again, $\lambda$ represents the variational parameters to be optimized. Since at test time we do not have access to $y$, we use the prior network and Monte Carlo sample in $p(y_* | x_*, \mathcal{D}) = \int p_\lambda(y_* | x_*, z) p_\lambda(z | x_*) \, \mathrm{d}z$. The specific form of the likelihood $p(y | x, z)$ can be found in the original paper \cite{kohl2018probabilistic}. This method is known to produce very diverse samples, from which we could estimate aleatoric uncertainty. In this paper, we endow the Probabilistic U-Net with a mechanism to estimate epistemic uncertainty and extend this method yet further, such that the aleatoric uncertainty estimates are automatically calibrated to the training set.

\section{Method} \label{sec:method}
We improve upon the Probabilistic U-Net model with two innovations. First, the original framework does not contain a mechanism to measure epistemic uncertainty. This can be included by adding variational dropout \cite{vdp} after the last convolution layer in the U-Net. This corresponds to setting $\theta=(z,w)$ and $q_\lambda(\theta)=q_\lambda(w)q_\lambda(z|x,y)$, where $w$ are CNN weights. The objective defined in Eq. \ref{eq:probunetloss} then changes to:
\begin{align}
    \begin{split}
        \mathcal{L_{\text{vd}}}(\lambda) = & -\frac{1}{N}\sum_{i=1}^N\E_{q_{\lambda}(w)q_\lambda(z| x_i, y_i)} \left [\log p_\lambda(y_i | x_i, z, w) \right ] + \\
        &\beta \cdot \frac{1}{N}\sum_{i=1}^N\infdiv{q_\lambda(z | x_i, y_i)}{p_\lambda(z | x_i)} + \frac{1}{N}\infdiv{q_{\lambda}(w)}{p(w)}. \label{eq:vd}
    \end{split}
\end{align}
Notice that as $N$ becomes very large the relative weight of the last KL term reduces, so the prior on $w$ is ignored \cite{galthesis}. The objective is maximized when $q_{\lambda}(w)$ is a delta peak on the maximum likelihood parameters, corresponding to zero model uncertainty. For our second innovation, we use intergrader variability $\VV_{p(\mathcal{D})}[y]$ as a training target for the predicted aleatoric uncertainty $\EE_{q_\lambda(\theta)}[\VV_{p(y|\theta,x)}[y]]$. We found directly minimizing the $L_1$ or $L_2$ distance between the two does not work well. Instead, since for binary variables the mean and variance are tied, we match the means $p_g = \EE_{p(\mathcal{D})}[y]$ and $p_m = \EE_{q_\lambda(\theta)}[\EE_{p(y|\theta,x)}[y]]$ using a cross-entropy loss. This term is not part of the ELBO, so we are free to sample from the prior network since this is used at test-time. Introducing scaling coefficient $\gamma$ our final training objective becomes
\begin{align}
    \mathcal{L}(\lambda) = \mathcal{L_{\text{vd}}}(\lambda) - \gamma \cdot \frac{1}{N}\sum_{i=1}^N [p_g \log p_m + (1-p_g) \log (1-p_m)]. \label{eq:trainobj2}
\end{align}

\section{Experiments}
\noindent\textbf{Datasets and Implementation Details}
We use two datasets where images have different but plausible annotations. First, the LIDC-IDRI dataset with 4 lesion annotations per image \cite{lidc1,lidc2,lidc3}. This dataset contains 1,018 lung CT scans from 1,010 lung patients with manual lesion annotations. We use the LIDC Matlab Toolbox \cite{toolbox} to process the slices and annotations with dimension 512 $\times$ 512, then center the lesions and crop the patches of size 128 $\times$ 128. This results in 15,096 image patches in total. We do not change the in-plane resolution. Second, the MICCAI2012 dataset with 3 prostate peripheral zone annotations per image \cite{da16}. This dataset contains 48 prostate MRI images and each image has multiple slides. We discard images that have fewer than 3 annotations, which leaves 44 images in total. For each image, the original dimension of a slide is 320 $\times$ 320, and we crop the central patch of size 128 $\times$ 128. This results in 614 image patches in total. The patches are treated independently and we feed each 2D patch to a model.  Since this is a very small dataset, we use elastic transformation \cite{elastrans} to augment the dataset to prevent overfitting.

\sloppy For each dataset, we split the train/validation/test sets with ratio 70\%/15\%/15\%. Different than \cite{kohl2018probabilistic}, we put all annotations of an image in the same mini-batch. Table \ref{tab:impdetails} shows some hyperparameters used for each dataset. The ones not presented in this table are similar to \cite{kohl2018probabilistic}. For ease of comparison, all experiments on the same dataset use the same hyperparameters. Lastly, we run all experiments on NVIDIA TITAN Xp GPUs.

\begin{table}[t]
\centering
\begin{tabular}{c|c|c}
Hyperparameters & LIDC  & MICCAI2012 \\ \hline
\# epochs & 800 & 1000 \\
Mini-batch size & 32 & 12 \\
$\beta$ & 1 & 100 \\
$\gamma$ & 100 & 100 \\
Data augmentation? & None & Elastic transformation \\
Adam learning rate & 1e-6 & 1e-4 
\end{tabular}
\caption{Hyperparameters details.}
\label{tab:impdetails}
\end{table}

\begin{figure*}
    \centering
        \begin{subfigure}[b]{0.475\textwidth}
        \centering
        \includegraphics[width=\textwidth]{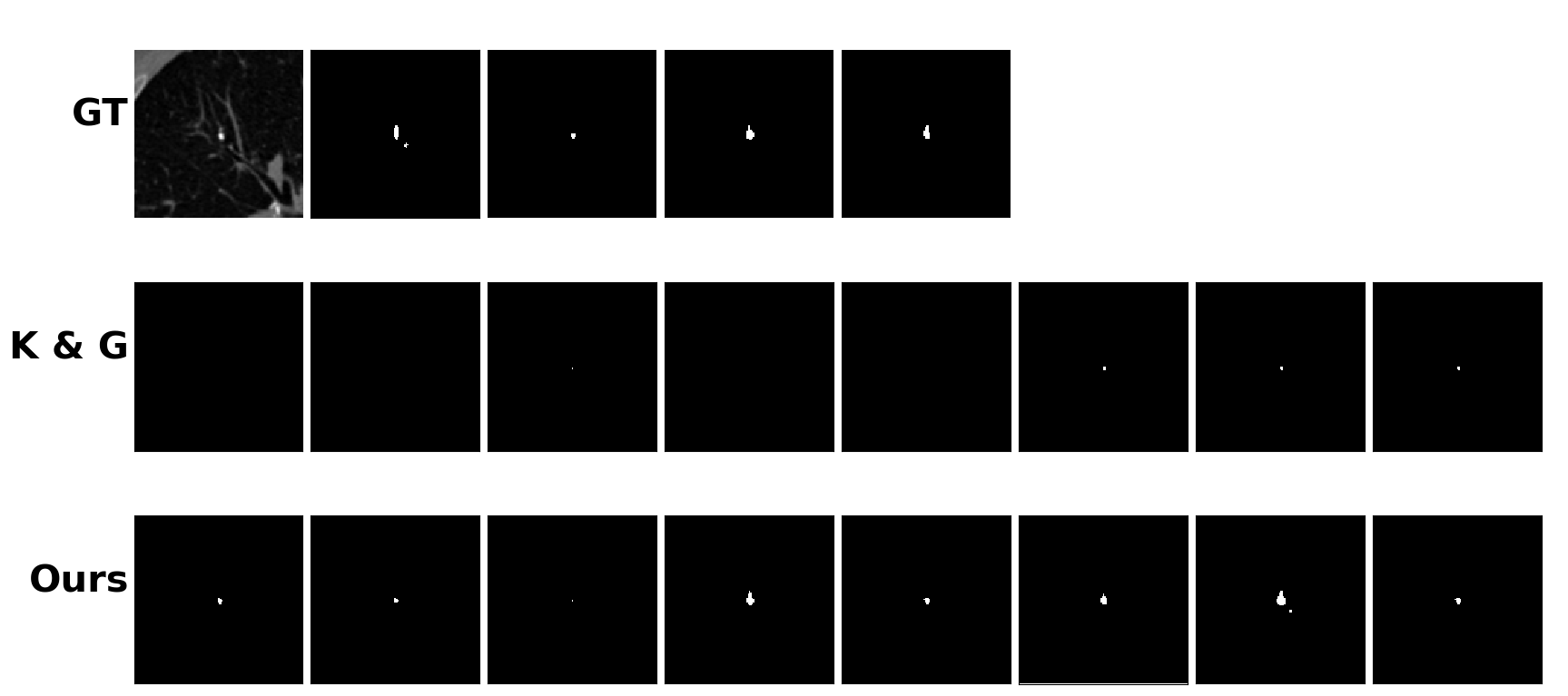}
            \caption{LIDC test sample 1}
            \label{fig:lidc1}
        \end{subfigure}
        \hfill
        \begin{subfigure}[b]{0.475\textwidth}  
            \centering 
            \includegraphics[width=\textwidth]{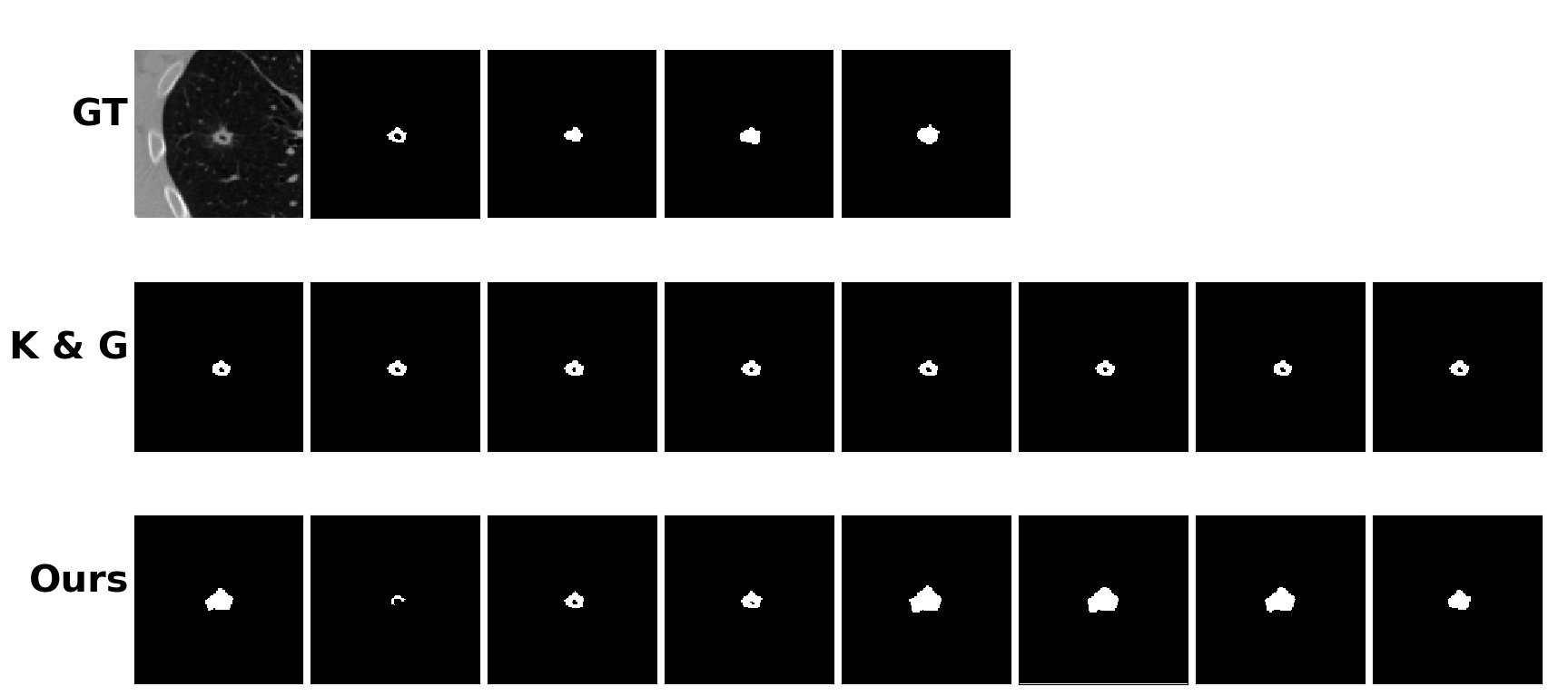}
            \caption{LIDC test sample 2}
            \label{fig:lidc2}
        \end{subfigure}
        \vskip\baselineskip
        \begin{subfigure}[b]{0.475\textwidth}   
            \centering 
            \includegraphics[width=\textwidth]{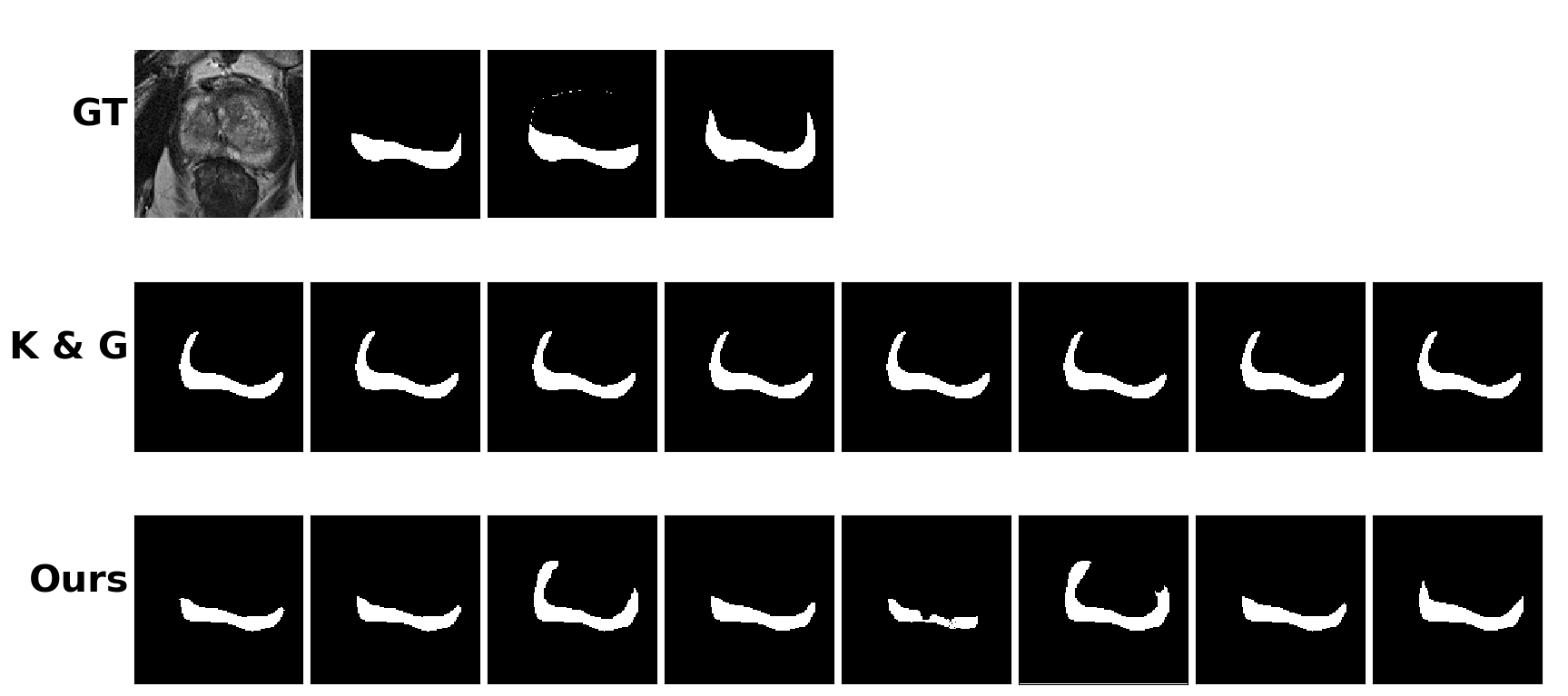}
            \caption{MICCAI2012 test sample 1}
            \label{fig:da16-1}
        \end{subfigure}
        \quad
        \begin{subfigure}[b]{0.475\textwidth}   
            \centering 
            \includegraphics[width=\textwidth]{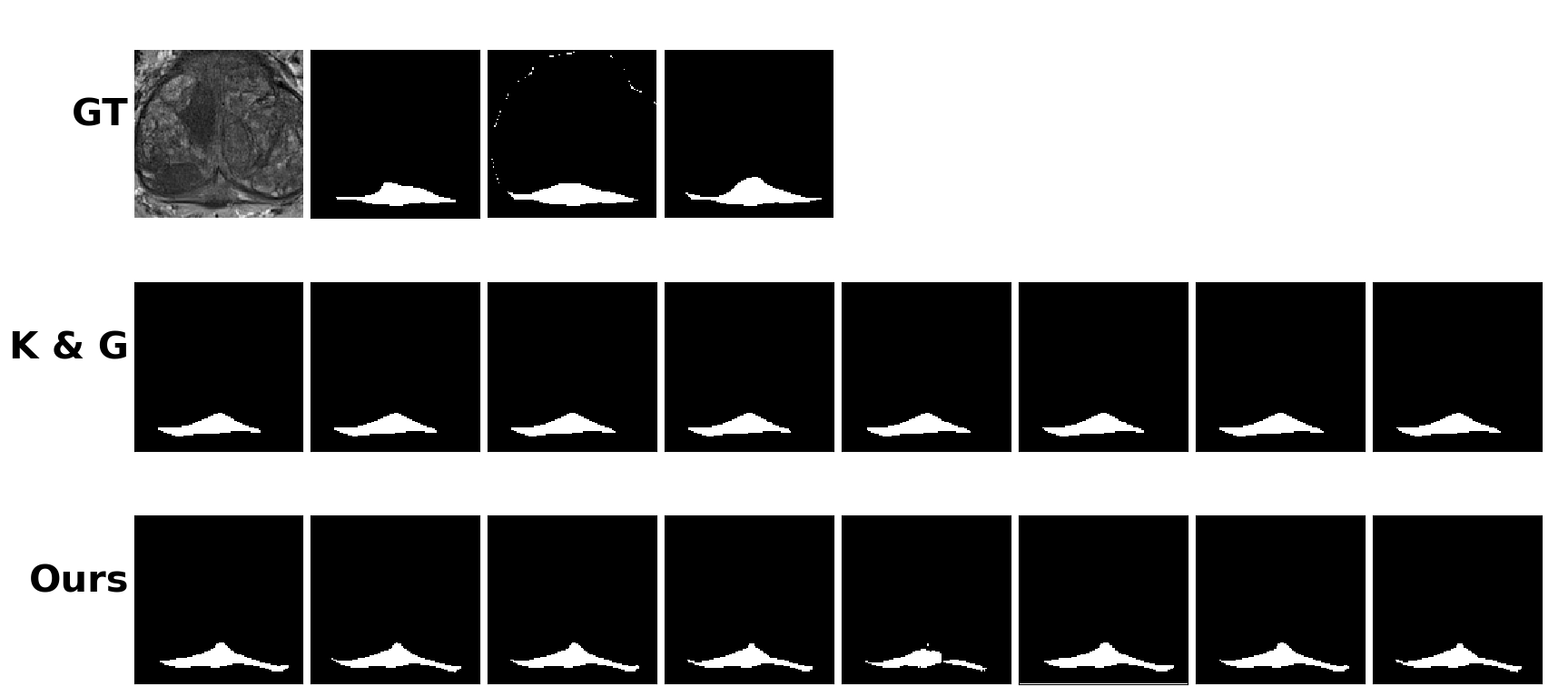}
            \caption{MICCAI2012 test sample 2}    
            \label{fig:da16-2}
        \end{subfigure}
        \caption
        {Samples comparison (zoom in on detail).} 
        \label{fig:samples}
\end{figure*}

\noindent\textbf{Sample Accuracy and Diversity}
Figure \ref{fig:samples} compares our generative results with Kendall and Gal model. In each plot, the first row is a patch in test and its true annotations. The second row is the samples generated by the Kendall and Gal model \cite{kendallgal}. The third row is our samples. The annotations for most images exhibit some variability as they come from different graders. In general, we observe that the samples from our model are able to cover different modality in the annotations, whereas in Kendall and Gal there is limited diversity, and thus cannot cover all the variations in the true annotations.

Quantitatively, we evaluate the generative results using the generalized energy distance (or $D^2_\text{GED}$) metric \cite{kohl2018probabilistic}: $2\E[d(S, Y)] - \E[d(S, S')] - \E[d(Y, Y')], \label{eq:ged}$
where $d$ is the complement of the Intersection over Union (IoU): $d(a, b) = 1 -  \text{IoU}(a, b)$. $S$ and $S'$ are independent samples from a model, and $Y$ and $Y'$ are independent samples from the graders. Thus, the first term measures the expected difference between the samples and annotations, the second among the samples themselves and the third among the annotations themselves. In other words, this metric evaluates both the accuracy and diversity of the samples.
Table \ref{tab:ged} compares the $D_{\text{GED}}^2$ scores on the LIDC and MICCAI2012 datasets. Since our model improves upon the Probabilistic U-Net model, we also present their numerical results for a reference\footnote{We use the PyTorch implementation for the Probabilistic U-Net model from \url{https://github.com/stefanknegt/Probabilistic-Unet-Pytorch}.}. Their generative results do not look very differently from ours, so we omit them in Figure \ref{fig:samples}. For each model in Table \ref{tab:ged}, we generate 50 samples for evaluation. The table shows our model achieves better $D_{\text{GED}}^2$ on both datasets. 
\begin{table}[t] 
\centering
\begin{tabular}{l|c|c}
 Method & LIDC & MICCAI2012 \\ \hline
 Kohl et al.  \cite{kohl2018probabilistic} & 0.346 $\pm$ 0.038 & 0.382 $\pm$ 0.017 \\
 Kendall \& Gal \cite{kendallgal} & 0.553 $\pm$ 0.010 & 0.571 $\pm$ 0.028 \\  
 Ours & \ubold{0.267 $\pm$ 0.012} & \ubold{0.373 $\pm$ 0.021}
\end{tabular}
\caption{$D_{\text{GED}}^2$ comparison (lower is better). Each result is computed over five random seeds.}
\label{tab:ged}
\end{table}

\begin{figure*}
    \centering
        \begin{subfigure}[b]{0.475\textwidth}
        \centering
        \includegraphics[width=\textwidth]{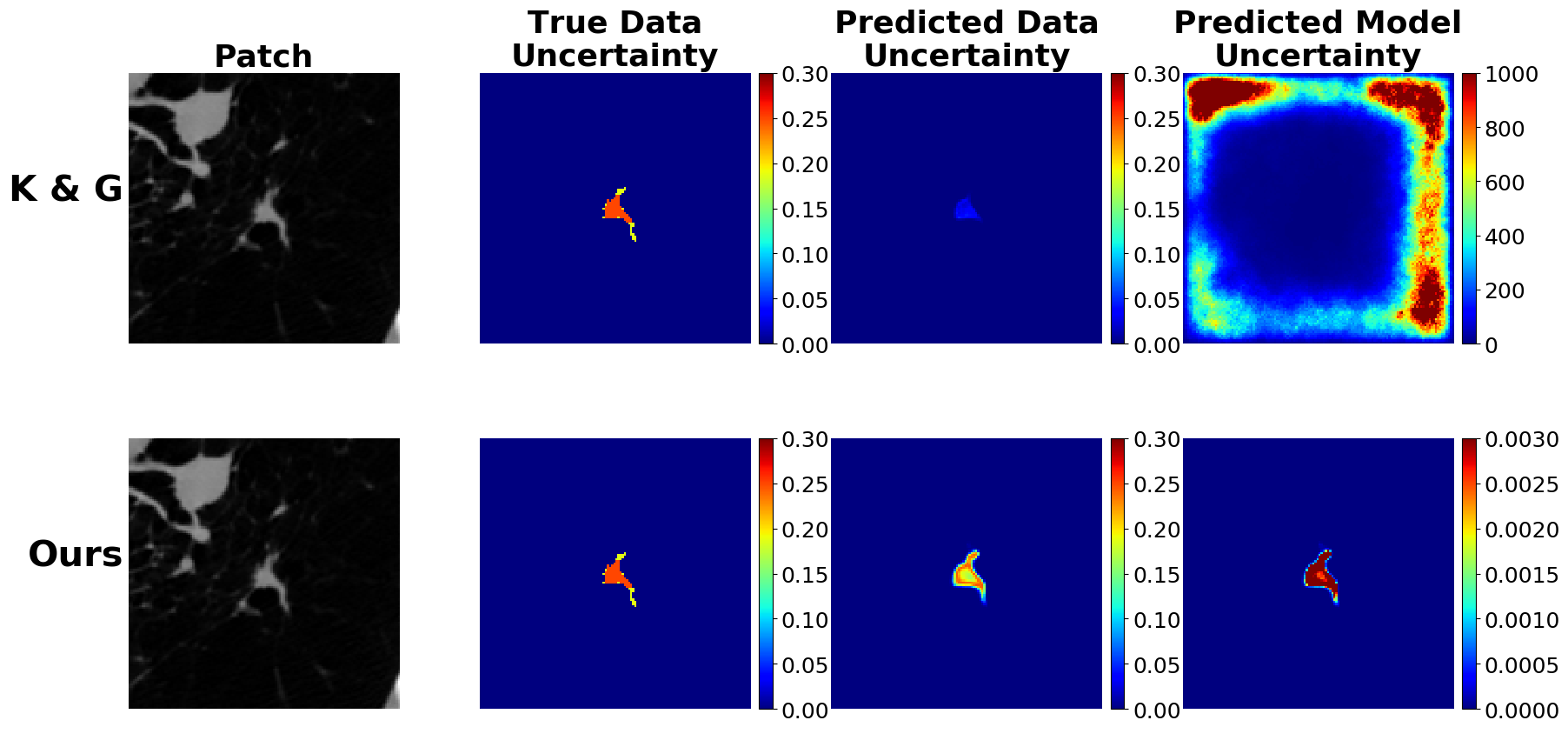}
            \caption{LIDC test sample 1}
            \label{fig:lidcunc1}
        \end{subfigure}
        \hfill
        \begin{subfigure}[b]{0.475\textwidth}  
            \centering 
            \includegraphics[width=\textwidth]{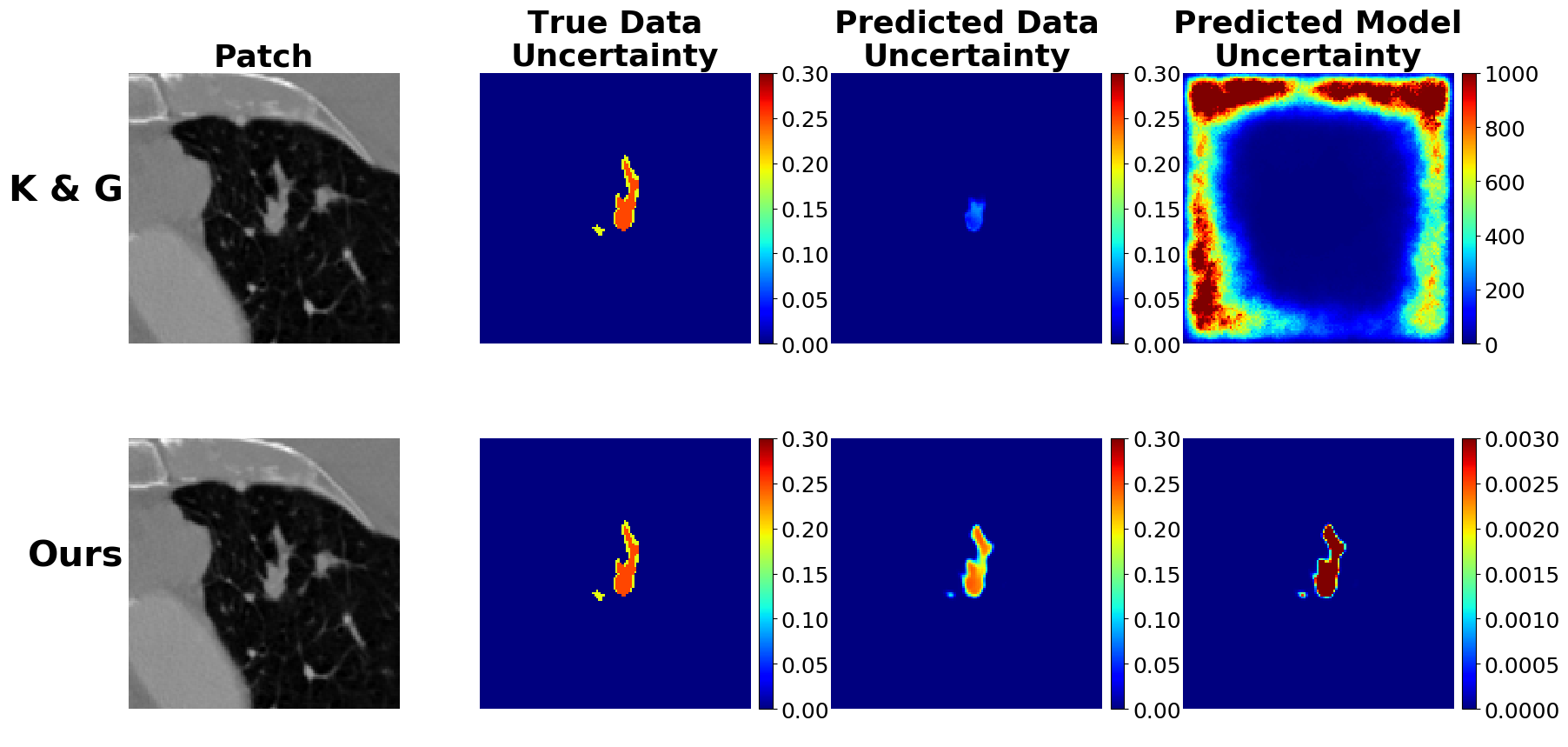}
            \caption{LIDC test sample 2}
            \label{fig:lidcunc2}
        \end{subfigure}
        \vskip\baselineskip
        \begin{subfigure}[b]{0.475\textwidth}   
            \centering 
            \includegraphics[width=\textwidth]{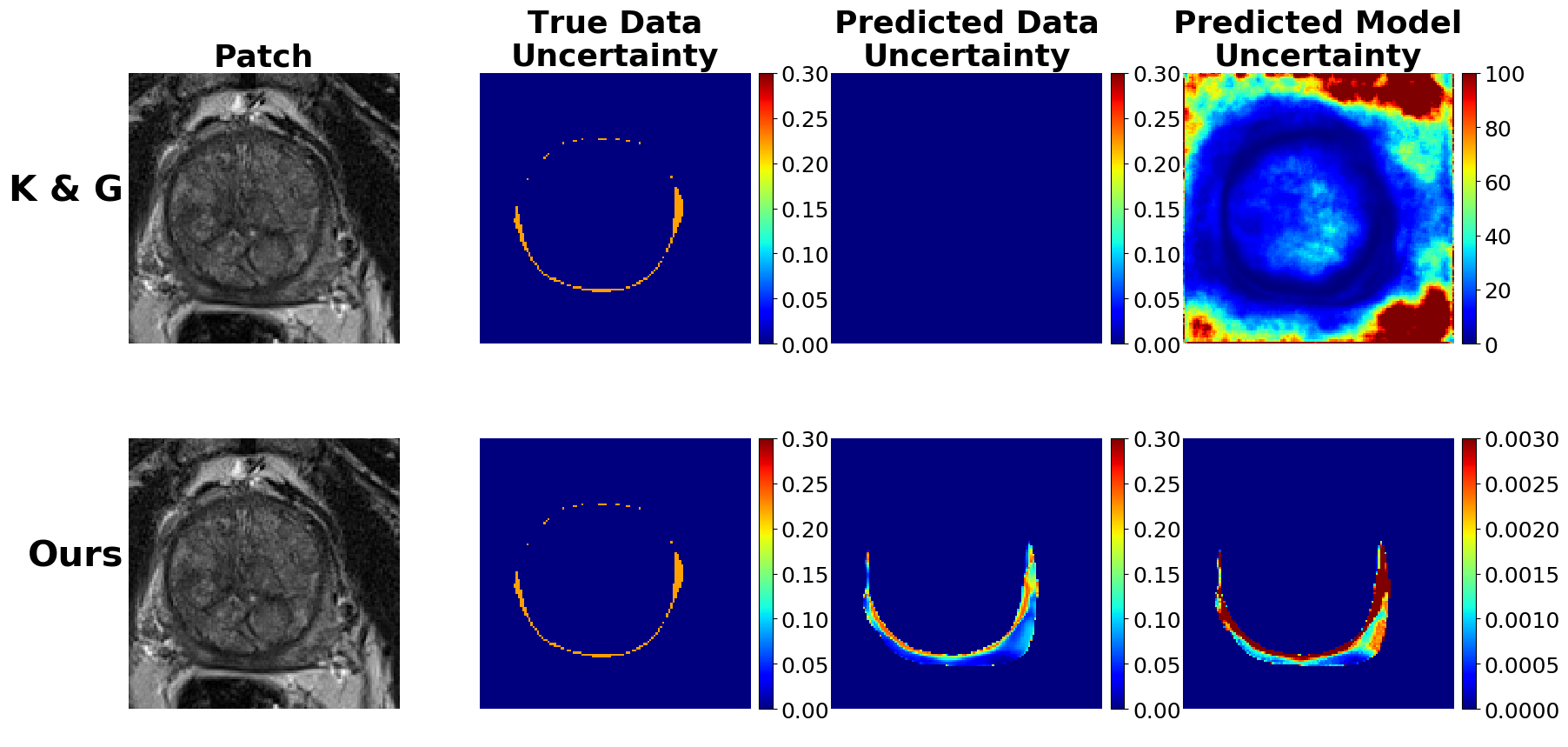}
            \caption{MICCAI2012 test sample 1}
            \label{fig:da16unc1}
        \end{subfigure}
        \quad
        \begin{subfigure}[b]{0.475\textwidth}   
            \centering 
            \includegraphics[width=\textwidth]{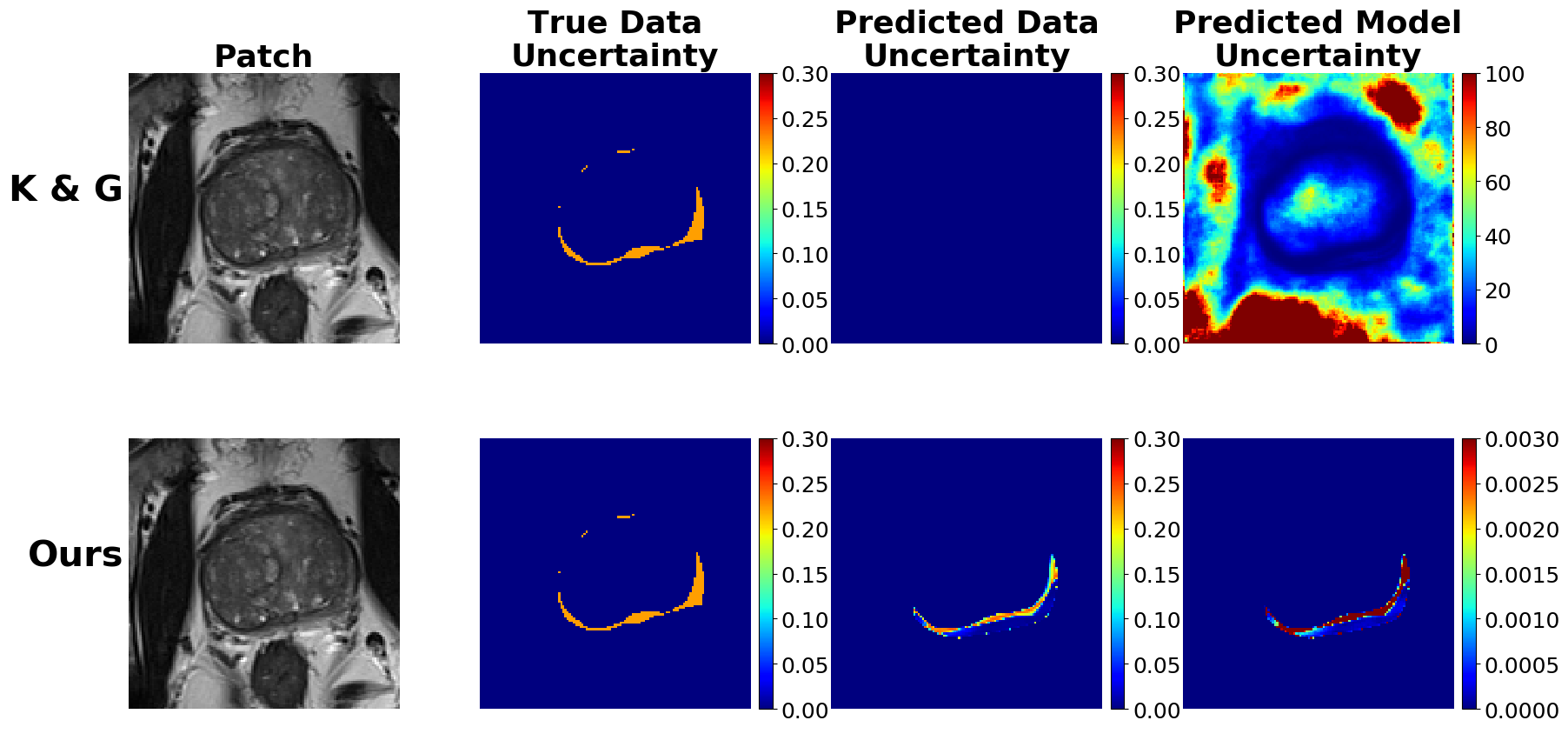}
            \caption{MICCAI2012 test sample 2}    
            \label{fig:da16unc2}
        \end{subfigure}
        \caption
        {Uncertainty quantification comparison.}
        \label{fig:unc}
\end{figure*}

\noindent\textbf{Uncertainty Decomposition} \label{sec:uncdecomp}
Figure \ref{fig:unc} shows the aleatoric and epistemic uncertainty decomposition results. For each plot, the first row shows the results from Kendall and Gal \cite{kendallgal} and the second row shows ours. To make the scales of these plots comparable, we set an upper threshold on the intensity values. Any value larger than the threshold will be treated as the threshold. For the true and predicted data uncertainty plots, we use the same threshold as we want to visually compare their similarity. In contrast, there is no label for the epistemic uncertainty, so we use a scale that fits well with most intensity pixels in a plot.

In general, we observe that Kendall and Gal makes plausible predictions on the shape of the data uncertainty, but tends to underestimate its scale, whereas we are relatively close to the ground truth in terms of both the shape and scale. Furthermore, the former tends to have high model uncertainty, especially at the image borders, whereas ours are usually around the point of interest. Although we do not know the true appearance of the model uncertainty, it should be high on the objects that do not occur often in the training set. In both test sets, the new objects usually appear around the center rather than at the borders. Therefore, we argue our epistemic uncertainty prediction is more sensible.

Quantitatively, Table \ref{tab:aleadiff} compares the data uncertainty prediction performance of the two models. As mentioned, the scale of the data uncertainty predictions from Kendall and Gal tends to be smaller than the ground truth. We want to establish a fair image similarity comparison that takes into account of this fact. Thus, for the true and predicted data uncertainty map $a$ and $b$, we measure their similarity using the normalized cross-correlation $\frac{1}{n \sigma_a \sigma_b} \sum_{x,y} (a_{x,y} - \mu_a) \cdot (b_{x,y} - \mu_b) \label{eq:ncc}$, where $n$ is the total number of pixels in an uncertainty map, and $\mu$ and $\sigma$ are the mean and the standard deviation of a uncertainty map. Since we normalize the scales of the uncertainty maps $a$ and $b$, the output value represents their intrinsic similarity. In Table \ref{tab:aleadiff}, we report the average normalized cross-correlation score over all test images for each dataset. Our model achieves higher data uncertainty correlations in both cases.
\begin{table}[t] 
\centering
\begin{tabular}{l|c|c}
 Method & LIDC & MICCAI2012 \\ \hline
 Kendall \& Gal \cite{kendallgal} & 0.597 $\pm$ 0.006 & 0.299 $\pm$ 0.011 \\ 
 Ours & \ubold{0.669 $\pm$ 0.011} & \ubold{0.345 $\pm$ 0.005}
\end{tabular}
\caption{Data uncertainty prediction comparison using the normalized cross-correlation (higher is better). Each result is computed over three random seeds.}
\label{tab:aleadiff}
\end{table}

\section{Conclusions}
In this work we designed a model for segmentation based on the Probabilistic U-Net \cite{kohl2018probabilistic} which outputs two kinds of quantifiable uncertainty, aleatoric (data) uncertainty and epistemic (model) uncertainty. We leveraged intergrader variability as a target for calibrated aleatoric uncertainty, which, as far as we know, related works have surprisingly not used. We showcased our model on the LIDC-IDRI lung nodule CT dataset \cite{lidc1,lidc2,lidc3} and MICCAI2012 prostate MRI dataset \cite{da16}, demonstrating that we could improve predictive uncertainty estimates. We also found that we could improve sample accuracy and sample diversity. In real-world applications, our method could inform doctors if they can trust the segmentation results to decide in real-time whether to continue a treatment path or not. As future work, we would like to improve the quality of the epistemic uncertainty.

\section*{Acknowledgements}
We thank Dimitrios Mavroeidis for helpful discussions and Arsenii Ashukha for the variational dropout code. This research was supported by NWO Perspective Grants DLMedIA and EDL, as well as the in-cash and in-kind contributions by Philips.

\bibliographystyle{splncs04}
\bibliography{mybibliography}

\end{document}